\documentclass[conference]{IEEEtran}
\IEEEoverridecommandlockouts
\usepackage{cite}
\usepackage{amsmath,amssymb,amsfonts}
\usepackage{algorithmic}
\usepackage{url}
\usepackage{graphicx}
\usepackage{textcomp}
\UseRawInputEncoding
\usepackage{xcolor}
\def\BibTeX{{\rm B\kern-.05em{\sc i\kern-.025em b}\kern-.08em
    T\kern-.1667em\lower.7ex\hbox{E}\kern-.125emX}}
    
\usepackage{eso-pic}
\newcommand\AtPageUpperMyright[5]{\AtPageUpperLeft{%
 \put(\LenToUnit{0.075\paperwidth},\LenToUnit{-1cm}){%
     \parbox{0.65\textwidth}{\raggedright\fontsize{9}{11}\selectfont #1}}%
 }}%
\newcommand{\conf}[1]{%
\AddToShipoutPictureBG*{%
\AtPageUpperMyright{#1}
}}

\begin{document}
\title{Social Robot Mediator for Multiparty Interaction}
\conf{2023 IEEE International Conference on Robotics and Automation (ICRA 2023) 
Workshop Towards a Balanced Cyberphysical Society: A Focus on Group Social Dynamics}
\author{\IEEEauthorblockN{Manith Adikari}
\IEEEauthorblockA{\textit{PhD Student} \\
\textit{Manchester Centre for Robotics and AI}\\
\textit{University of Manchester}\\
\textit{manith.adikari@postgrad.manchester.ac.uk} \\
\and
\IEEEauthorblockN{Angelo Cangelosi}
\IEEEauthorblockA{\textit{Professor of Machine Learning and Robotics} \\
\textit{Manchester Centre for Robotics and AI}\\
\textit{University of Manchester}\\
\textit{angelo.cangelosi@manchester.ac.uk}\\
\and
\IEEEauthorblockN{Randy Gomez}
\IEEEauthorblockA{\textit{Chief Scientist} \\
\textit{Honda Research Institute Japan} \\
\textit{Tokyo, Japan} \\
\textit{r.gomez@jp.honda-ri.com}\\
}}}}

\maketitle
\thispagestyle{plain}
\pagestyle{plain}
\begin{abstract}
A social robot acting as a ‘mediator’ can enhance interactions between humans, for example, in fields such as education and healthcare. A particularly promising area of research is the use of a social robot mediator in a multiparty setting, which tends to be the most applicable in real-world scenarios. However, research in social robot mediation for multiparty interactions is still emerging and faces numerous challenges. This paper provides an overview of social robotics and mediation research by highlighting relevant literature and some of the ongoing problems. The importance of incorporating relevant psychological principles for developing social robot mediators is also presented. Additionally, the potential of implementing a Theory of Mind in a social robot mediator is explored, given how such a framework could greatly improve mediation by reading the individual and group mental states to interact effectively. 
\end{abstract}
\hfill \break

\section{Introduction}
\subsection{Human-Robot Interaction}

Human-Robot Interaction (HRI) is a diverse and multidisciplinary field which notably appeared in the mid-1990s and has continued to grow over the last few decades. HRI can be defined as the ‘field of study dedicated to understanding, designing, and evaluating robotic systems for use by or with humans’ \cite{goodrichHumanRobotInteractionSurvey2007}. Currently, HRI is applied in numerous applications, from industrial automation \cite{hentoutHumanRobotInteraction2019} to elderly care \cite{abdiScopingReviewUse2018}. Additionally, modern HRI research highlights the value of a ‘multidisciplinary approach’ involving robotics, psychology, cognitive science and other relevant research. With the rapid advancements of robotic technologies, in recent years, there has also been an emphasis on developing HRI with a greater awareness of the human and societal impacts \cite{lasotaSurveyMethodsSafe2017}. For instance, a study outlined how robotic technologies can help achieve the United Nations (UN) Sustainable Development Goals (SDGs) \cite{maiRoleRoboticsAchieving2022a}.

\subsection{Social Robotics}
Social Robotics are an emerging technology which is increasingly being implemented in real-world applications, particularly in healthcare as Socially Assistive Robots (SARs) and education \cite{abdiScopingReviewUse2018,belpaemeSocialRobotsEducation2018}. The exact concept and definition of a ‘Social Robot’ has varied in literature: ‘socially evocative’, ‘sociable’, and ‘socially intelligent’ \cite{breazealEmotionSociableHumanoid2003, fongSurveySociallyInteractive2003, dautenhahnSociallyIntelligentRobots2007a}. The survey \cite{fongSurveySociallyInteractive2003} proposed the term ‘socially interactive robot’ referring to a robot whose social interaction plays a key role in HRI, as opposed to robots which have ‘conventional’ HRI like teleoperated industrial robots. In general, current literature tends to follow this interpretation to describe social robots, though there is no universally accepted definition \cite{dautenhahnSociallyIntelligentRobots2007, belpaemeSocialRobotsEducation2018}.  

Since the development of the earliest social robots, such as MIT’s Kismet in the 1990s, robots have evolved from being mere tools to social collaborators and assistants.  With these advancements, social robots have the potential to promote human-human interactions as social ‘mediators’ in areas such as education and healthcare \cite{rietherSocialFacilitationSocial2012,gilletSocialRobotMediator2020,shimadaHowCanSocial2012}. Current research has shown the effectiveness of social robot mediators in improving task participation \cite{gilletRobotGazeCan2021a,tennentMicbotPeripheralRobotic2019}, fostering inclusion \cite{10.1145/3568162.3576997,strohkorbseboStrategiesInclusionHuman2020}, and positively impacting group dynamics \cite{shimadaHowCanSocial2012}. The area of social robot mediator research is still developing and will be a key focus of this paper.

\begin{figure}[htbp]
\centerline{\includegraphics[scale=0.52]{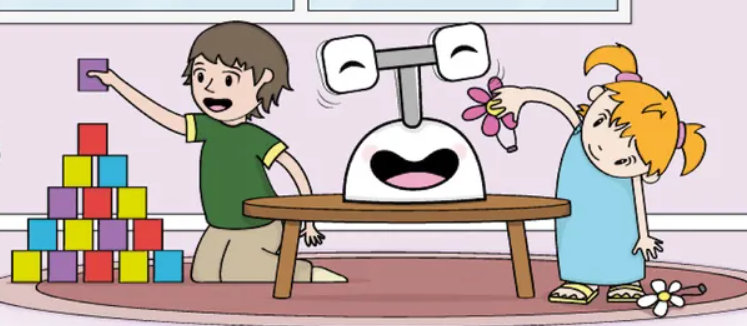}}
\caption{An example of a multiparty interaction with Haru the social robot from Honda Research Institute \cite{ieeeMeetHaru}.}
\label{fig1}
\end{figure}

Currently, most social robotics literature focuses on ‘One-to-One’ human-robot interactions, where one human participant interacts with one robot (often referred to as 'dyadic interactions'). However, in recent years, there has been a shift to research ‘multiparty’ human-robot interactions (also referred to as ‘non-dyadic’ interactions). Multiparty social interactions are prevalent in real-world settings, such as classrooms and hospitals, hence the increasing research focus on this topic. Developing social robots capable of multiparty interaction will be necessary to utilize robots in a wider range of real-world applications. However, there are additional research challenges which need to be overcome. This paper focuses on multiparty interaction with a social robot mediator in a ‘Many-to-One’ scenario, where multiple human participants interact with one robot - for example, shown in Figure \ref{fig1}.

In some studies, a social mediator robot is also applied in the context of resolving conflict. However, this paper will focus on social mediation in ‘harmonious’ situations, where social interaction is facilitated, rather than conflict. Specifically, a social robot mediator in a multiparty setting (‘Many-to-One’ scenario) to facilitate social interaction.

Over the past few decades, the social robotics market has experienced substantial growth and is expected to continue growing significantly in the future \cite{imarcgroupSocialRobots}. Despite this, numerous social robotic ventures for real-world applications have ended in failure \cite{parviainenPoliticalChoreographySophia2021}. Though the failures can be attributed to numerous factors, such as insufficient hardware and high costs, a major reason is the difficulty of developing effective human-robot social interaction. Although addressing this multifaceted research challenge is complex, a promising direction is to firmly incorporate cognitive and psychological principles \cite{koppFabricSociallyInteractive2022, Cangelosi_Schlesinger_2015, lungarellaDevelopmentalRoboticsSurvey2003a}. This paper will present the potential of involving cognitive science and psychology theories, as well as a ‘Theory of Mind’ framework. Overall, this paper proposes the novelty of a social robot mediator in a multiparty setting utilizing a Theory of Mind framework. 

\section{Social Robot Mediator}

As social robotics research continues to advance, a social robot ‘mediator’ could enhance current state-of-the-art human-robot interaction and be valuable for real-world applications. In general, human social mediators could facilitate human interactions in various social settings, such as school classrooms and formal meetings. However, it often becomes impractical and challenging to involve a human mediator – for instance, due to cost and logistical reasons. This signifies the necessity and value of using a robot social mediator; for example, a social robot could be used as a tool to mediate groups of school children \cite{gilletSocialRobotMediator2020}.

The specific definition of a social mediator differs through the literature across different fields and use cases. Vygotsky \cite{Vygotskiĭ_Hanfmann_Kozulin_Vakar_2013} proposed that a mediator can be any resource to help individuals learn or problem-solve; mediators can be objects, symbols (for example, language) or even a social entity (in this case, a social robot). The concept of a robot ‘mediator’ in current literature is generally consistent with the previously mentioned definition \cite{gilletSocialRobotMediator2020, matsuyamaFourparticipantGroupConversation2015}. It is also important to note that ‘facilitation’ is when the presence of another (for instance, a social robot) can affect the performance during a task (for example, collaborating between two groups); this is in line with the definition in social psychology \cite{Allport_Drury_Hare_Borgatta_Bales_1955}. 

Given the broad scope of social mediation research, relevant work has been done not just within social robotics, but also with other ‘Socially Interactive Agents’ (SIAs). SIAs could be defined as ‘virtually or physically embodied agents that are capable of autonomously communicating with people and each other in a socially intelligent manner using multi-modal behaviours’ \cite[p.~1]{lugrinIntroductionSociallyInteractive2021}. The broad term of SIAs encompasses technologies such as Social Robots, Intelligent Virtual Agents (IVAs) and Embodied Conversational Agents (ECAs). Social mediation and multiparty setting research in these fields are relevant and will therefore be briefly examined.  

\subsection{Virtually Embodied Agents}

SIAs encompass virtually embodied agents, like IVAs and ECAs, as well as physically embodied agents, like social robots. The following section will specifically explore social mediation in multiparty settings using virtually embodied SIAs, which are embodied agents in a virtual environment capable of social interaction, like digital assistants such as Apple’s Siri and Amazon’s Alexa.

As previously mentioned, a social mediator can positively influence human interactions in practical settings. However, it is not always possible to have a human mediator for practical reasons. To address these, digital technologies such as SIAs have been recognized as powerful tools to aid communication, in this case through social mediation. Most research in virtually embodied SIAs is based on individual human interactions, such as personal assistants and tutors \cite{lugrinIntroductionSociallyInteractive2021}. Recently, there is emerging research in using virtual social mediators for multiparty settings, which presents additional challenges, as well as greater opportunities for practical applications. Several research areas are examined such as the perception and understanding of group dynamics as well as generating group behaviours like conversation \cite{gilletMultipartyInteractionHumans2022}. Broadly, these studies present insights into social behaviours, multimodal perception methods, applicable computational models and paths for future research relevant to social robotics. However, there are obvious differences between virtual agents compared to a physically embodied agents, such as the operation in a more complex real-world environment. Therefore, it is important to consider the applicability of these findings within the framework of a social robot.

\subsection{Social Robot: A Physically Embodied Agent}

\begin{figure}[htbp]
\centerline{\includegraphics[scale=0.35]{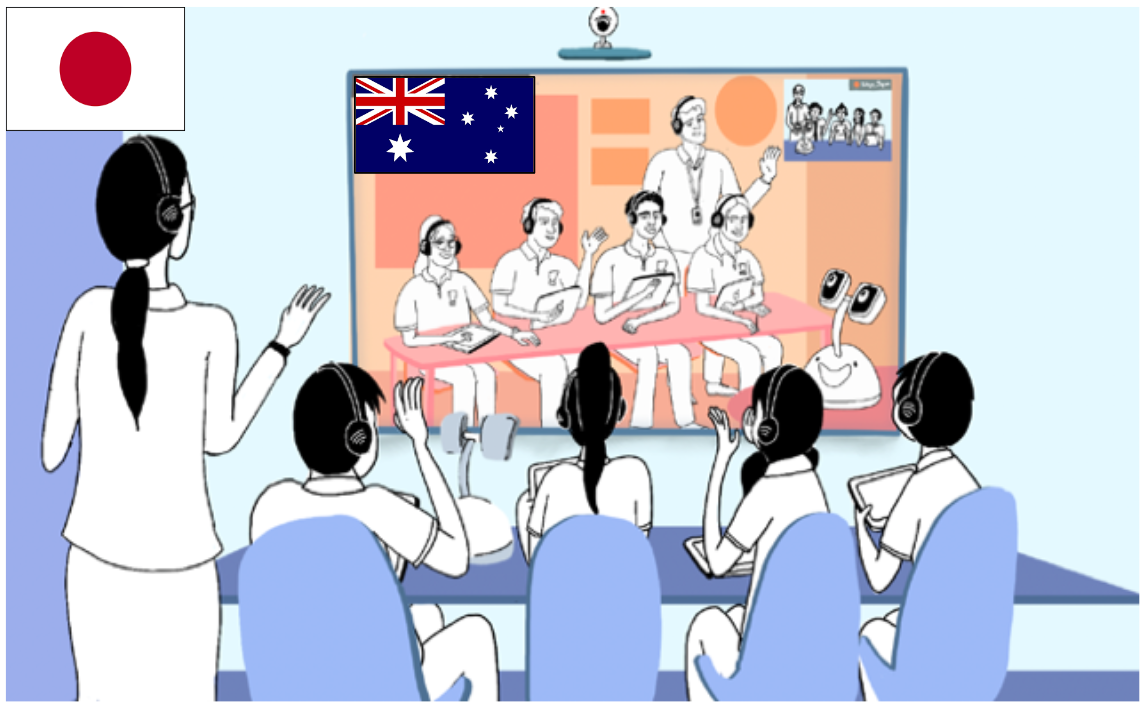}}
\caption{Set-up for Haru social robot as a classroom mediator between a class in Japan and Australia \cite[Figure.~4]{mypersonalrobotsEmbodiedMediator}}
\label{fig2}
\end{figure}

Social robots acting as mediators have been effective at improving human interaction through numerous aspects, such as facilitating conversation, engagement, task collaboration, participation, inclusion, and even trustworthiness \cite{strohkorbseboStrategiesInclusionHuman2020}. A practical scenario of a social robot mediator is shown in Figure \ref{fig2}.

A significant benefit of a social robot mediator compared to virtual SIAs is the difference in embodiment. The definition of ‘embodiment’ can vary depending on the field of study, whether it is philosophy or psychology, but in social robotics, the term ‘physical embodiment’ refers to the ‘physical instantiation of an agent in its environment, adapted into the concept of “physical grounding”’ \cite{dengref}. The benefit of physical embodiment has been previously explored in psychology; for instance, the Media Equation theory \cite{lee2004presence} mentions how the physical presence of an agent (in this case, a social robot) could improve social interaction. A social robot, being physically embodied, can be significantly impactful in enhancing engagement, trustworthiness, collaboration and communication \cite{dengref}. Most of the social robotics literature focuses on robots working for humans (for example, as assistants) rather than alongside humans (in this case, as mediators) \cite{aylettUnsocialRobotsHow2023}. However, there is emerging research in social robot mediators particularly for multiparty settings, which is the primary focus of this paper. 

Over the past decades there has been more research in social mediation in multiparty interactions for numerous use-cases such as fostering inclusion \cite{strohkorbseboStrategiesInclusionHuman2020, gilletSocialRobotMediator2020}, facilitating conversation \cite{matsuyamaFourparticipantGroupConversation2015}, improving collaboration \cite{shimadaHowCanSocial2012}, and balancing participation \cite{tennentMicbotPeripheralRobotic2019,zarkowskiMultipartyTurnTakingRepeated2019}.

The literature also delves into how different social behaviours could affect mediation: gaze \cite{mutluFootingHumanRobotConversations,gilletRobotGazeCan2021}, gestures \cite{kondoGestureCentricAndroidSystem2013}, supportive or vulnerable expressions \cite{traegerVulnerableRobotsPositively2020,strohkorbseboStrategiesInclusionHuman2020}.  There are other studies which investigate social elements such as touch, embodiment, proxemics, and turn-taking \cite{gilletMultipartyInteractionHumans2022,zarkowskiMultipartyTurnTakingRepeated2019}.

In general, the social robot architectures within these studies tend to be multimodal with incremental processing across multiple directions between components (‘Multi-directional, incremental architectures’); an overview of existing social robot architectures is given in \cite{koppFabricSociallyInteractive2022}. However, these architectures often do not have a strong basis in psychology or cognitive science principles, which is important to consider when developing effective social robots.

\section{Relevant Psychology Principles}
Incorporating psychology principles into social robotics research has been noted as a promising path to address the research challenges \cite{goodrichHumanRobotInteractionSurvey2007,fongSurveySociallyInteractive2003,lungarellaDevelopmentalRoboticsSurvey2003}. Notably, this contributed to the formation of the ‘Developmental Robotics’ field, which applies developmental psychology in robotics research \cite{Cangelosi_Schlesinger_2015, asadaCognitiveDevelopmentalRobotics2009}. However, a significant portion of current social robotics literature tends not to have a firm basis in such cognitive and psychology theories \cite{koppFabricSociallyInteractive2022}. In the context of social robot mediators in multiparty settings, there are several key areas to consider, as mentioned below. 

\begin{figure}[htbp]
\centerline{\includegraphics[scale=0.8]{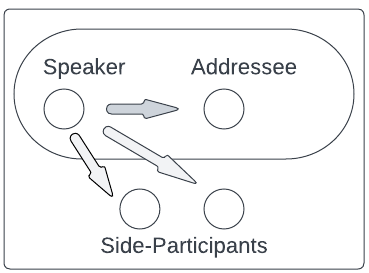}}
\caption{Participant structure of a multiparty interaction \cite{Goffman_2010a,Clark_2009}}
\label{fig3}
\end{figure}

Goffman \cite{Goffman_2010a} and Clark \cite{Clark_2009} presented the participant structure of a multiparty interaction, as shown in Figure \ref{fig3}. There are participants in the interaction and then there are non-participants. The participants could be divided into three roles: Speaker, the primary speaker; Addressee, the primary listener; Side-Participant, who is not the addressed listener. Moreover, throughout the multiparty interaction, the participants continuously change roles thereby progressing the social interaction. Latane presented the psychology of ‘social impact,’ which describes how individuals are impacted in a multiparty setting \cite{Lat81}.

It is also useful to examine the earlier research which defines key terms like ‘social interaction’ \cite{Goffman_2010a} and 'jointly focused gathering' \cite{Kendon_2009}. Other psychology theories relevant to multiparty settings should also be considered, such as ‘small group’ dynamics \cite{Bales_2013}, ‘proxemics’ \cite{Kendon_2009}, and engagement \cite{sidner2005explorations}.

There are studies in social robot mediation which do incorporate these theories: \cite{matsuyamaFourparticipantGroupConversation2015,bohusFacilitatingMultipartyDialog2010,mutluFootingHumanRobotConversations}. 

\section{Theory of Mind}
Another concept which has seldom been applied to social robot mediator research yet holds significant potential to improve social interaction is ‘Theory of Mind’ (ToM). ToM is the ability to ‘infer thoughts, feelings, and beliefs of others’\cite{leslie1987}. ToM has been proven to develop from a young age in humans and be important for social interaction \cite{yottinfant}. This enables humans to recognize others can have different goals, beliefs, desires, and emotions to themselves. Within a social robot mediator, a ToM could enable reading the mental states of a multiparty interaction and then responding appropriately. 

Since ToM was first proposed by Premack in the 1970s \cite{premackDoesChimpanzeeHave1978}, further work has been done to conceptualize a ToM model as well as implement in robots – most notably \cite{baron-cohenMindblindnessEssayAutism,leslieToMMToBYAgency1994,scassellatiFoundationsTheoryMind2001}. The success of human interactions has been attributed to the capabilities of humans to model cognitive processes and the state-of-mind of other humans, thus improving decision-making in social interactions \cite{patricioMathematicalModelsTheory2022,scassellatiFoundationsTheoryMind2001}. Implementing ToM in robots have shown to improve HRI, for instance, by increasing perception of anthropomorphism thereby improving trust or by simply enabling robots to realise the underlying intentions of human agents to facilitate task collaboration \cite{romeoExploringTheoryMind2022, vinanziWouldRobotTrust2019}. The presence of ToM is commonly tested in humans and robots using false belief tasks, which test the ability to recognize that others can have different beliefs to themselves \cite{Wimmer1983}. The ‘Sally-Anne’ task \cite{BaronCohen1985} is the most common variation of the original false belief task which is used in current literature. 

Currently, one of the biggest challenges in robotics is to develop proficient social robots, capable of autonomously inferring mental states (for example, intentions and beliefs) of people across different scenarios and environments \cite{yangGrandChallengesScience2018}. In addition to implementing the general social psychology principles mentioned above, a ToM framework could address some of the existing challenges in HRI, as well as social robotics. At present, it is not clearly defined how humans exhibit ToM which has led to various approaches in psychology and cognitive science \cite{inbook}. Computational modelling of ToM typically uses either ‘Association’, ‘Teleological’, or ‘Simulation’ theories \cite{biancoPsychologicalIntentionRecognition2020, biancoTransferringAdaptiveTheory2019}. Up-and-coming research has implemented ToM primarily for perspective-taking within interactions and tasks, enabling robots to realize different mental states (e.g. goals) of people, thereby improving interaction. Additionally, ToM has been used in combination with machine learning-based methods for purposes such as learning about real-world environments, usually identifying behaviours of agents and various object dynamics \cite{Vanderelst2018}. In the case of a social robot mediator, a ToM framework could greatly improve the mediation process by reading the intentions of different individuals and groups as well as responding accordingly in order to achieve better social interaction. Despite this, there have been no studies specifically using ToM for social mediation in multiparty interactions.

Artificial Intelligence (AI) advancements have led to promising ToM implementations based on AI methods. However, researchers in robotics and AI utilize only parts of ToM psychology and cognitive science research to propose models specifically for robotics applications without necessarily representing human ToM. This leads to different approaches to model ToM in robotics which can be categorized as ‘Cognitive architecture’, ‘Decision theoretic’, or ‘Deep learning’ implementations \cite{gurneyRobotsTheoryMind2022}. Decision theoretic approaches, such as Partially Observable Markov Decision Processes (POMDPs), have proven to be promising given how flexible and interpretable they are. However, there is emerging work in using deep neural networks to model ToM in robots, however, there are difficulties in developing interpretable models based on social psychology \cite{gurneyRobotsTheoryMind2022}.

\section{Challenges \& Ideas for Designing Robot Mediators}
A social robot mediator could improve numerous social interactions, for instance by facilitating collaboration, inclusion or conversation. In multiparty situations, there is an even greater opportunity for robot mediators to positively impact real-world settings, such as education and healthcare. This is a goal of many social robotics researchers which is yet to be fully realized due to the numerous research challenges in this field. The following section discusses the current challenges of designing social robot mediators and potential solutions to address them. This will entail drawing insights from current research accomplishments and expanding upon them. 

\subsection{Real world HRI testing}
This paper previously presented the promise of applying general psychology and cognitive science principles to develop social robots, as well as the potential of a ToM framework. It would be valuable to also consider specific psychology theories depending on the context and interaction, for example, mediation strategies in a school classroom setting with children. This would justify the design decisions of social robot platforms and allow for more generalizable research studies. Moreover, it would be beneficial to validate the research through real-world human interactions, potentially related to a practical scenario. The benefits of real-world as well as long-term, repeated HRI have been noted frequently \cite{lungarellaDevelopmentalRoboticsSurvey2003}, however, this is scarce in multiparty social robotics literature. Conducting such experiments would truly demonstrate the effectiveness of the research as well as highlight the difficulties of practical implementation, which will be important to consider if robots are to be utilized in real-world settings. For generalizability in future work, it would be useful to consider shortcomings in current HRI experiments, such as limited statistical power and highly specific methodologies \cite{limSocialRobotsGlobal2021}. Concerning social mediation, it would be valuable to examine the practical applications of a social robot mediator. A potential use case is using a robot mediator in education environments like a school classroom to encourage social interaction between students. For example, it can be to promote learning or inclusion through cross-cultural or interdisciplinary interactions. This would explore interesting research paths within multiparty HRI, such as intra-group/intergroup social mediation and interaction in diverse or dynamic settings. Furthermore, multiparty HRI research is not yet fully established, so more research could be done regarding metrics and replicability \cite{gilletMultipartyInteractionHumans2022}.

\subsection{Theory of Mind implemented for practical scenarios}
A social robot mediator in a multiparty setting would greatly benefit from a ToM framework since it would enable intention reading of individuals and mediate appropriately depending on the scenario, much like how humans do. Additionally, it would be promising to explore how reading not just individual intentions but also group intentions using a ToM approach could enhance mediation. A compelling use case could involve enabling mediation between diverse groups, such as those from different cultures or disciplinary backgrounds. However, given there is no consensus on how humans attempt ToM, the exact meaning of implementing ToM can be subjective \cite{gurneyRobotsTheoryMind2022}. Therefore, as mentioned in \cite{gurneyRobotsTheoryMind2022}, it is crucial to define the ToM implementation within the study and outline what it entails. Additionally, the addition of a multiparty interaction introduces further complexity and interest to current ToM implementation in robotics. Notably, it will be valuable to explore how relevant psychology theories and principles interconnect with the ToM framework. Future work could also explore how social robot mediators with ToMs behave across diverse scenarios, potentially mimicking real-world settings, which would show the efficacies of different ToM models and help evaluate them. It would also be interesting to develop robots exhibiting more ‘active’ social behaviours using a ToM to make them more useful participants during interactions. For instance, a robot could ‘actively perceive’ social cues and then ‘proactively’ interact with humans \cite{biancoFunctionalAdvantagesAdaptive2019}. In addition to current ToM research related to improving HRI trust, it would also be worthwhile to explore how a ToM could facilitate trustworthiness between human-human interactions in a multiparty setting – for example, between student groups of different cultures. Developing social robots based on emerging AI methods could be promising, such as using deep learning based ToM frameworks or even taking inspiration from various domains like Natural Language Generation (NLG) \cite{Foster2019}. In the case of social mediation, drawing insights from relevant research like ‘AI-mediated communication’ \cite{10.1093/jcmc/zmz022} could be promising. 

\subsection{Considering societal impacts \& ethics}
Finally, when developing social robots, particularly for real-world implementation, it would be important to consider the wider societal impacts and ethics. For example, when designing a social robot mediator for a classroom setting, it would be crucial to ensure the mediation itself is necessary and beneficial as well as ensuring it does not hinder diversity or inclusion. Though this can be subjective, as robotics research advances it would be necessary to consider these subjects if robots are to be truly utilized practically. Examples of such emerging research include the Haksh-E robot \cite{hatrob} and Haru social robot \cite{aylettUnsocialRobotsHow2023} which consider design principles from relevant organizations like UNICEF and WHO.

\section{Conclusion}
A social robot mediator could facilitate social interactions between humans, enabling better engagement and collaboration. Such a mediator is especially valuable in multiparty settings, which are ubiquitous in the real world. However, it is still an ongoing research problem to develop social robots capable of perceiving, rationalizing, and reacting effectively to mediate. Research in social robot mediation is still growing and could greatly benefit psychology and cognitive science theories mentioned previously. Also, it is valuable to have a ToM framework as it could facilitate social interaction by appropriately reading and handling the mental states of different humans in a multiparty setting. Since, most social robotics research focuses on individual HRI rather than multiparty settings, further research in this area will explore interesting problems like reading group intentions using ToM or applying group psychology principles to mediate. Hence, this area of research in social robot mediation in multiparty settings using ToM is a novel and promising avenue likely to be addressed in future work. 

\bibliographystyle{ieeetr}
\bibliography{biblib}

\begin{thebibliography}{10}

\bibitem{goodrichHumanRobotInteractionSurvey2007}
M.~A. Goodrich and A.~C. Schultz, ``Human-{Robot} {Interaction}: {A}
  {Survey},'' {\em Foundations and Trends® in Human-Computer Interaction},
  vol.~1, no.~3, pp.~203--275, 2007.

\bibitem{hentoutHumanRobotInteraction2019}
A.~Hentout, M.~Aouache, A.~Maoudj, and I.~Akli, ``Human–robot interaction in
  industrial collaborative robotics: a literature review of the decade
  2008–2017,'' {\em Advanced Robotics}, vol.~33, pp.~764--799, Aug. 2019.

\bibitem{abdiScopingReviewUse2018}
J.~Abdi, A.~Al-Hindawi, T.~Ng, and M.~P. Vizcaychipi, ``Scoping review on the
  use of socially assistive robot technology in elderly care,'' {\em BMJ Open},
  vol.~8, p.~e018815, Feb. 2018.

\bibitem{lasotaSurveyMethodsSafe2017}
P.~A. Lasota, T.~Fong, and J.~A. Shah, ``A {Survey} of {Methods} for {Safe}
  {Human}-{Robot} {Interaction},'' {\em Foundations and Trends in Robotics},
  vol.~5, no.~3, pp.~261--349, 2017.

\bibitem{maiRoleRoboticsAchieving2022a}
V.~Mai, B.~Vanderborght, T.~Haidegger, A.~Khamis, N.~Bhargava, D.~B. Boesl,
  K.~Gabriels, A.~Jacobs, A.~Moon, R.~Murphy, Y.~Nakauchi, E.~Prestes,
  B.~Rao~R., R.~Vinuesa, and C.-M. Morch, ``The {Role} of {Robotics} in
  {Achieving} the {United} {Nations} {Sustainable} {Development}
  {Goals}—{The} {Experts}' {Meeting} at the 2021 {IEEE}/{RSJ} {IROS}
  {Workshop} [{Industry} {Activities}],'' {\em IEEE Robotics \& Automation
  Magazine}, vol.~29, pp.~92--107, Mar. 2022.

\bibitem{belpaemeSocialRobotsEducation2018}
T.~Belpaeme, J.~Kennedy, A.~Ramachandran, B.~Scassellati, and F.~Tanaka,
  ``Social robots for education: {A} review,'' {\em Science Robotics}, vol.~3,
  p.~eaat5954, Aug. 2018.

\bibitem{breazealEmotionSociableHumanoid2003}
C.~Breazeal, ``Emotion and sociable humanoid robots,'' {\em International
  Journal of Human-Computer Studies}, vol.~59, pp.~119--155, July 2003.

\bibitem{fongSurveySociallyInteractive2003}
T.~Fong, I.~Nourbakhsh, and K.~Dautenhahn, ``A survey of socially interactive
  robots,'' {\em Robotics and Autonomous Systems}, vol.~42, pp.~143--166, Mar.
  2003.

\bibitem{dautenhahnSociallyIntelligentRobots2007a}
K.~Dautenhahn, ``Socially intelligent robots: dimensions of human–robot
  interaction,'' {\em Philosophical Transactions of the Royal Society B:
  Biological Sciences}, vol.~362, pp.~679--704, Apr. 2007.

\bibitem{dautenhahnSociallyIntelligentRobots2007}
K.~Dautenhahn, ``Socially intelligent robots: dimensions of human–robot
  interaction,'' {\em Philosophical Transactions of the Royal Society B:
  Biological Sciences}, vol.~362, pp.~679--704, Apr. 2007.

\bibitem{rietherSocialFacilitationSocial2012}
N.~Riether, F.~Hegel, B.~Wrede, and G.~Horstmann, ``Social facilitation with
  social robots?,'' in {\em Proceedings of the seventh annual {ACM}/{IEEE}
  international conference on {Human}-{Robot} {Interaction}}, (Boston
  Massachusetts USA), pp.~41--48, ACM, Mar. 2012.

\bibitem{gilletSocialRobotMediator2020}
S.~Gillet, W.~van~den Bos, and I.~Leite, ``A social robot mediator to foster
  collaboration and inclusion among children,'' in {\em Robotics: {Science} and
  {Systems} {XVI}}, Robotics: Science and Systems Foundation, July 2020.

\bibitem{shimadaHowCanSocial2012}
M.~Shimada, T.~Kanda, and S.~Koizumi, ``How {Can} a {Social} {Robot}
  {Facilitate} {Children}'s {Collaboration}?,'' in {\em Social {Robotics}}
  (D.~Hutchison, T.~Kanade, J.~Kittler, J.~M. Kleinberg, F.~Mattern, J.~C.
  Mitchell, M.~Naor, O.~Nierstrasz, C.~Pandu~Rangan, B.~Steffen, M.~Sudan,
  D.~Terzopoulos, D.~Tygar, M.~Y. Vardi, G.~Weikum, S.~S. Ge, O.~Khatib, J.-J.
  Cabibihan, R.~Simmons, and M.-A. Williams, eds.), vol.~7621, pp.~98--107,
  Berlin, Heidelberg: Springer Berlin Heidelberg, 2012.
\newblock Series Title: Lecture Notes in Computer Science.

\bibitem{gilletRobotGazeCan2021a}
S.~Gillet, R.~Cumbal, A.~Pereira, J.~Lopes, O.~Engwall, and I.~Leite, ``Robot
  {Gaze} {Can} {Mediate} {Participation} {Imbalance} in {Groups} with
  {Different} {Skill} {Levels},'' in {\em Proceedings of the 2021 {ACM}/{IEEE}
  {International} {Conference} on {Human}-{Robot} {Interaction}}, (Boulder CO
  USA), pp.~303--311, ACM, Mar. 2021.

\bibitem{tennentMicbotPeripheralRobotic2019}
H.~Tennent, S.~Shen, and M.~Jung, ``Micbot: {A} {Peripheral} {Robotic} {Object}
  to {Shape} {Conversational} {Dynamics} and {Team} {Performance},'' in {\em
  2019 14th {ACM}/{IEEE} {International} {Conference} on {Human}-{Robot}
  {Interaction} ({HRI})}, (Daegu, Korea (South)), pp.~133--142, IEEE, Mar.
  2019.

\bibitem{10.1145/3568162.3576997}
I.~Neto, F.~Correia, F.~Rocha, P.~Piedade, A.~Paiva, and H.~Nicolau, ``The
  robot made us hear each other: Fostering inclusive conversations among
  mixed-visual ability children,'' in {\em Proceedings of the 2023 ACM/IEEE
  International Conference on Human-Robot Interaction}, HRI '23, (New York, NY,
  USA), p.~13–23, Association for Computing Machinery, 2023.

\bibitem{strohkorbseboStrategiesInclusionHuman2020}
S.~Strohkorb~Sebo, L.~L. Dong, N.~Chang, and B.~Scassellati, ``Strategies for
  the {Inclusion} of {Human} {Members} within {Human}-{Robot} {Teams},'' in
  {\em Proceedings of the 2020 {ACM}/{IEEE} {International} {Conference} on
  {Human}-{Robot} {Interaction}}, (Cambridge United Kingdom), pp.~309--317,
  ACM, Mar. 2020.

\bibitem{ieeeMeetHaru}
https://www.facebook.com/48576411181, ``Meet haru, the unassuming big-eyed
  robot helping researchers study social robotics --- spectrum.ieee.org.''
  \url{https://spectrum.ieee.org/honda-research-institute-haru-social-robot}.
\newblock [Accessed 11-May-2023].

\bibitem{imarcgroupSocialRobots}
``{S}ocial {R}obots {M}arket {S}ize, {S}hare, {I}ndustry {T}rends {R}eport
  2023-2028 --- imarcgroup.com.''
  \url{https://www.imarcgroup.com/social-robots-market }.
\newblock [Accessed 11-May-2023].

\bibitem{parviainenPoliticalChoreographySophia2021}
J.~Parviainen and M.~Coeckelbergh, ``The political choreography of the {Sophia}
  robot: beyond robot rights and citizenship to political performances for the
  social robotics market,'' {\em AI \& SOCIETY}, vol.~36, pp.~715--724, Sept.
  2021.

\bibitem{koppFabricSociallyInteractive2022}
S.~Kopp and T.~Hassan, ``The {Fabric} of {Socially} {Interactive} {Agents}:
  {Multimodal} {Interaction} {Architectures},'' in {\em The {Handbook} on
  {Socially} {Interactive} {Agents}} (B.~Lugrin, C.~Pelachaud, and D.~Traum,
  eds.), pp.~77--112, New York, NY, USA: ACM, 1~ed., Oct. 2022.

\bibitem{Cangelosi_Schlesinger_2015}
A.~Cangelosi and M.~Schlesinger, {\em Developmental robotics: From babies to
  robots}.
\newblock The MIT Press, 2015.

\bibitem{lungarellaDevelopmentalRoboticsSurvey2003a}
M.~Lungarella, G.~Metta, R.~Pfeifer, and G.~Sandini, ``Developmental robotics:
  a survey,'' {\em Connection Science}, vol.~15, pp.~151--190, Dec. 2003.

\bibitem{matsuyamaFourparticipantGroupConversation2015}
Y.~Matsuyama, I.~Akiba, S.~Fujie, and T.~Kobayashi, ``Four-participant group
  conversation: {A} facilitation robot controlling engagement density as the
  fourth participant,'' {\em Computer Speech \& Language}, vol.~33, pp.~1--24,
  Sept. 2015.

\bibitem{Allport_Drury_Hare_Borgatta_Bales_1955}
F.~H. Allport, J.~Drury, A.~P. Hare, E.~F. Borgatta, and R.~F. Bales, {\em The
  influence of the group upon association and thought}.
\newblock 1955.

\bibitem{lugrinIntroductionSociallyInteractive2021}
B.~Lugrin, ``Introduction to {Socially} {Interactive} {Agents},'' in {\em The
  {Handbook} on {Socially} {Interactive} {Agents}} (B.~Lugrin, C.~Pelachaud,
  and D.~Traum, eds.), pp.~1--20, New York, NY, USA: ACM, 1~ed., Sept. 2021.

\bibitem{gilletMultipartyInteractionHumans2022}
S.~Gillet, M.~Vázquez, C.~Peters, F.~Yang, and I.~Leite, ``Multiparty
  {Interaction} {Between} {Humans} and {Socially} {Interactive} {Agents},'' in
  {\em The {Handbook} on {Socially} {Interactive} {Agents}} (B.~Lugrin,
  C.~Pelachaud, and D.~Traum, eds.), pp.~113--154, New York, NY, USA: ACM,
  1~ed., Oct. 2022.

\bibitem{mypersonalrobotsEmbodiedMediator}
``Embodied mediator for cross cultural understanding among children across the
  world — socially intelligent robotics consortium ---
  mypersonalrobots.org.'' \url{https://mypersonalrobots.org/pilot}.
\newblock [Accessed 11-May-2023].

\bibitem{dengref}
E.~Deng, B.~Mutlu, and M.~J. Mataric, {\em Embodiment in Socially Interactive
  Robots}.
\newblock 2019.

\bibitem{lee2004presence}
K.~M. Lee, ``Why presence occurs: Evolutionary psychology, media equation, and
  presence,'' {\em Presence: Teleoperators \& Virtual Environments}, vol.~13,
  no.~4, pp.~494--505, 2004.

\bibitem{aylettUnsocialRobotsHow2023}
M.~P. Aylett, R.~Gomez, E.~Sandry, and S.~Sabanovic, ``Unsocial {Robots}: {How}
  {Western} {Culture} {Dooms} {Consumer} {Social} {Robots} to a {Society} of
  {One},'' in {\em Extended {Abstracts} of the 2023 {CHI} {Conference} on
  {Human} {Factors} in {Computing} {Systems}}, (Hamburg Germany), pp.~1--6,
  ACM, Apr. 2023.

\bibitem{zarkowskiMultipartyTurnTakingRepeated2019}
M.~Żarkowski, ``Multi-party {Turn}-{Taking} in {Repeated} {Human}–{Robot}
  {Interactions}: {An} {Interdisciplinary} {Evaluation},'' {\em International
  Journal of Social Robotics}, vol.~11, pp.~693--707, Dec. 2019.

\bibitem{mutluFootingHumanRobotConversations}
B.~Mutlu, T.~Shiwa, T.~Kanda, H.~Ishiguro, and N.~Hagita, ``Footing {In}
  {Human}-{Robot} {Conversations}: {How} {Robots} {Might} {Shape} {Participant}
  {Roles} {Using} {Gaze} {Cues},''

\bibitem{gilletRobotGazeCan2021}
S.~Gillet, R.~Cumbal, A.~Pereira, J.~Lopes, O.~Engwall, and I.~Leite, ``Robot
  {Gaze} {Can} {Mediate} {Participation} {Imbalance} in {Groups} with
  {Different} {Skill} {Levels},'' in {\em Proceedings of the 2021 {ACM}/{IEEE}
  {International} {Conference} on {Human}-{Robot} {Interaction}}, (Boulder CO
  USA), pp.~303--311, ACM, Mar. 2021.

\bibitem{kondoGestureCentricAndroidSystem2013}
Y.~Kondo, K.~Takemura, J.~Takamatsu, and T.~Ogasawara, ``A {Gesture}-{Centric}
  {Android} {System} for {Multi}-{Party} {Human}-{Robot} {Interaction},'' {\em
  Journal of Human-Robot Interaction}, vol.~2, pp.~133--151, Mar. 2013.

\bibitem{traegerVulnerableRobotsPositively2020}
M.~L. Traeger, S.~Strohkorb~Sebo, M.~Jung, B.~Scassellati, and N.~A.
  Christakis, ``Vulnerable robots positively shape human conversational
  dynamics in a human–robot team,'' {\em Proceedings of the National Academy
  of Sciences}, vol.~117, pp.~6370--6375, Mar. 2020.

\bibitem{lungarellaDevelopmentalRoboticsSurvey2003}
M.~Lungarella, G.~Metta, R.~Pfeifer, and G.~Sandini, ``Developmental robotics:
  a survey,'' {\em Connection Science}, vol.~15, pp.~151--190, Dec. 2003.

\bibitem{asadaCognitiveDevelopmentalRobotics2009}
M.~Asada, K.~Hosoda, Y.~Kuniyoshi, H.~Ishiguro, T.~Inui, Y.~Yoshikawa,
  M.~Ogino, and C.~Yoshida, ``Cognitive {Developmental} {Robotics}: {A}
  {Survey},'' {\em IEEE Transactions on Autonomous Mental Development}, vol.~1,
  pp.~12--34, May 2009.

\bibitem{Goffman_2010a}
E.~Goffman, {\em Forms of talk}.
\newblock University of Pennsylvania Press, 2010.

\bibitem{Clark_2009}
H.~H. Clark, {\em Using language}.
\newblock Cambridge University Press, 2009.

\bibitem{Lat81}
B.~Lantané, ``{The psychology of social impact},'' {\em American
  Psychologist}, vol.~36, pp.~343--356, 1981.

\bibitem{Kendon_2009}
A.~Kendon, {\em Conducting interaction: Patterns of behavior in focused
  encounters}.
\newblock Cambridge University Press, 2009.

\bibitem{Bales_2013}
R.~F. Bales, {\em Interaction process analysis a method for the study of small
  groups}.
\newblock Isha Books, 2013.

\bibitem{sidner2005explorations}
C.~L. Sidner, C.~Lee, C.~Kidd, N.~Lesh, and C.~Rich, ``Explorations in
  engagement for humans and robots,'' 2005.

\bibitem{bohusFacilitatingMultipartyDialog2010}
D.~Bohus and E.~Horvitz, ``Facilitating multiparty dialog with gaze, gesture,
  and speech,'' in {\em International {Conference} on {Multimodal} {Interfaces}
  and the {Workshop} on {Machine} {Learning} for {Multimodal} {Interaction}},
  (Beijing China), pp.~1--8, ACM, Nov. 2010.

\bibitem{leslie1987}
A.~Leslie, ``Pretense and representation: The origins of "theory of mind",''
  {\em Psychological Review}, vol.~94, pp.~412--426, 10 1987.

\bibitem{yottinfant}
J.~Yott and D.~Poulin-Dubois, ``Are infants' theory of mind abilities well
  integrated? implicit understanding of intentions, desires, and beliefs,''
  {\em Journal of Cognition and Development}, 02 2016.

\bibitem{premackDoesChimpanzeeHave1978}
D.~Premack and G.~Woodruff, ``Does the chimpanzee have a theory of mind?,''
  {\em Behavioral and Brain Sciences}, vol.~1, pp.~515--526, Dec. 1978.

\bibitem{baron-cohenMindblindnessEssayAutism}
S.~Baron-Cohen, {\em Mindblindness: an {Essay} on {Autism} and {Theory} of
  {Mind}}.

\bibitem{leslieToMMToBYAgency1994}
A.~M. Leslie, ``{ToMM}, {ToBY}, and {Agency}: {Core} architecture and domain
  specificity,'' in {\em Mapping the {Mind}} (L.~A. Hirschfeld and S.~A.
  Gelman, eds.), pp.~119--148, Cambridge University Press, 1~ed., Apr. 1994.

\bibitem{scassellatiFoundationsTheoryMind2001}
B.~M. Scassellati, ``Foundations for a theory of mind for a humanoid robot:
  (446982006-001),'' 2001.

\bibitem{patricioMathematicalModelsTheory2022}
M.~M. Patrício and A.~Jamshidnejad, ``Mathematical {Models} of {Theory} of
  {Mind},'' Sept. 2022.
\newblock arXiv:2209.14450 [cs].

\bibitem{romeoExploringTheoryMind2022}
M.~Romeo, P.~E. McKenna, D.~A. Robb, G.~Rajendran, B.~Nesset, A.~Cangelosi, and
  H.~Hastie, ``Exploring {Theory} of {Mind} for {Human}-{Robot}
  {Collaboration},'' in {\em 2022 31st {IEEE} {International} {Conference} on
  {Robot} and {Human} {Interactive} {Communication} ({RO}-{MAN})}, (Napoli,
  Italy), pp.~461--468, IEEE, Aug. 2022.

\bibitem{vinanziWouldRobotTrust2019}
S.~Vinanzi, M.~Patacchiola, A.~Chella, and A.~Cangelosi, ``Would a robot trust
  you? {Developmental} robotics model of trust and theory of mind,'' {\em
  Philosophical Transactions of the Royal Society B: Biological Sciences},
  vol.~374, p.~20180032, Apr. 2019.

\bibitem{Wimmer1983}
H.~Wimmer, ``Beliefs about beliefs: Representation and constraining function of
  wrong beliefs in young children{\textquotesingle}s understanding of
  deception,'' {\em Cognition}, vol.~13, pp.~103--128, Jan. 1983.

\bibitem{BaronCohen1985}
S.~Baron-Cohen, A.~M. Leslie, and U.~Frith, ``Does the autistic child have a
  {\textquotedblleft}theory of mind{\textquotedblright} ?,'' {\em Cognition},
  vol.~21, pp.~37--46, Oct. 1985.

\bibitem{yangGrandChallengesScience2018}
G.-Z. Yang, J.~Bellingham, P.~E. Dupont, P.~Fischer, L.~Floridi, R.~Full,
  N.~Jacobstein, V.~Kumar, M.~McNutt, R.~Merrifield, B.~J. Nelson,
  B.~Scassellati, M.~Taddeo, R.~Taylor, M.~Veloso, Z.~L. Wang, and R.~Wood,
  ``The grand challenges of \textit{{Science} {Robotics}},'' {\em Science
  Robotics}, vol.~3, p.~eaar7650, Jan. 2018.

\bibitem{inbook}
N.~Gurney, S.~Marsella, V.~Ustun, and D.~Pynadath, {\em Operationalizing
  Theories of Theory of Mind: A Survey}, pp.~3--20.
\newblock 01 2023.

\bibitem{biancoPsychologicalIntentionRecognition2020}
F.~Bianco and D.~Ognibene, ``From {Psychological} {Intention} {Recognition}
  {Theories} to {Adaptive} {Theory} of {Mind} for {Robots}: {Computational}
  {Models},'' in {\em Companion of the 2020 {ACM}/{IEEE} {International}
  {Conference} on {Human}-{Robot} {Interaction}}, (Cambridge United Kingdom),
  pp.~136--138, ACM, Mar. 2020.

\bibitem{biancoTransferringAdaptiveTheory2019}
F.~Bianco and D.~Ognibene, ``Transferring {Adaptive} {Theory} of {Mind} to
  {Social} {Robots}: {Insights} from {Developmental} {Psychology} to
  {Robotics},'' in {\em Social {Robotics}} (M.~A. Salichs, S.~S. Ge, E.~I.
  Barakova, J.-J. Cabibihan, A.~R. Wagner, Ã.~Castro-González, and H.~He,
  eds.), vol.~11876, pp.~77--87, Cham: Springer International Publishing, 2019.
\newblock Series Title: Lecture Notes in Computer Science.

\bibitem{Vanderelst2018}
D.~Vanderelst and A.~Winfield, ``An architecture for ethical robots inspired by
  the simulation theory of cognition,'' {\em Cognitive Systems Research},
  vol.~48, pp.~56--66, May 2018.

\bibitem{gurneyRobotsTheoryMind2022}
N.~Gurney and D.~V. Pynadath, ``Robots with {Theory} of {Mind} for {Humans}:
  {A} {Survey},'' in {\em 2022 31st {IEEE} {International} {Conference} on
  {Robot} and {Human} {Interactive} {Communication} ({RO}-{MAN})}, (Napoli,
  Italy), pp.~993--1000, IEEE, Aug. 2022.

\bibitem{limSocialRobotsGlobal2021}
V.~Lim, M.~Rooksby, and E.~S. Cross, ``Social {Robots} on a {Global} {Stage}:
  {Establishing} a {Role} for {Culture} {During} {Human}–{Robot}
  {Interaction},'' {\em International Journal of Social Robotics}, vol.~13,
  pp.~1307--1333, Sept. 2021.

\bibitem{biancoFunctionalAdvantagesAdaptive2019}
F.~Bianco and D.~Ognibene, ``Functional advantages of an adaptive {Theory} of
  {Mind} for robotics: a review of current architectures,'' in {\em 2019 11th
  {Computer} {Science} and {Electronic} {Engineering} ({CEEC})}, (Colchester,
  United Kingdom), pp.~139--143, IEEE, Sept. 2019.

\bibitem{Foster2019}
M.~E. Foster, ``Natural language generation for social robotics: opportunities
  and challenges,'' {\em Philosophical Transactions of the Royal Society B:
  Biological Sciences}, vol.~374, p.~20180027, Mar. 2019.

\bibitem{10.1093/jcmc/zmz022}
J.~T. Hancock, M.~Naaman, and K.~Levy, ``{AI-Mediated Communication:
  Definition, Research Agenda, and Ethical Considerations},'' {\em Journal of
  Computer-Mediated Communication}, vol.~25, pp.~89--100, 01 2020.

\bibitem{hatrob}
S.~Sasidharan, D.~Pasupuleti, A.~Das, C.~Kapoor, G.~Manikutty, and B.~Rao,
  ``Haksh-e: An autonomous social robot for promoting good hand hygiene among
  children,'' 10 2021.

\end{thebibliography}

\end{document}